%% file: main.tex
\documentclass{article}
\usepackage{authblk}
\usepackage[utf8]{inputenc}
\usepackage{main}
\usepackage{microtype}
\usepackage{subcaption}
\usepackage{graphicx}
\usepackage{times}
\usepackage{latexsym}
\usepackage{amsmath}
\usepackage{float}
\usepackage{footnote}
\usepackage{enumitem}
\usepackage{bm}
\usepackage{arydshln}
\usepackage{booktabs}
\usepackage{array}     
\usepackage{multicol}
\usepackage{multirow}
\usepackage{color}
\usepackage{xcolor}     
\usepackage{colortbl}
\usepackage{bbding}
\usepackage{makecell}
\usepackage{mathtools}
\usepackage{imakeidx}
\usepackage{longtable}
\usepackage{wrapfig}
\usepackage{algorithmic}
\usepackage{rotating}
\makeindex
\usepackage{arydshln}
\usepackage{lipsum}
\usepackage{natbib}
\usepackage[toc]{multitoc}
\usepackage[edges]{forest}
\usepackage[normalem]{ulem}
\definecolor{mydarkblue}{rgb}{0,0.08,0.45}
\usepackage[colorlinks=true,linkcolor=mydarkblue,citecolor=mydarkblue,filecolor=mydarkblue,urlcolor=mydarkblue]{hyperref}
\usepackage{CJKutf8}
\usepackage{awesomebox} 
\usepackage{bbding}
\usepackage[most]{tcolorbox}
\usepackage{booktabs}
\usepackage{geometry}
\definecolor{wkblue}{rgb}{0.2, 0.3, 0.6}
\definecolor{meta-color}{rgb}{0.5, 0.5, 0.5}
\usepackage{amsmath}
\usepackage{enumitem}
\usepackage{lscape} 
\usepackage{booktabs}
\usepackage{algorithm}   
\usepackage{setspace}

\usepackage{algorithmic}
\usepackage{tabularx,booktabs}
\usepackage{makecell}

\usepackage{amssymb}
\usepackage{amsfonts}

\usepackage{booktabs} 
\usepackage{geometry} 

\usepackage{multirow}

\usepackage[tikz]{bclogo}
\usepackage[framemethod=tikz]{mdframed}
\definecolor{bgblue}{RGB}{245,243,253}
\definecolor{ttblue}{RGB}{91,194,224}

\usepackage{pgfplots}
\usepackage{pgfplotstable}

\usepackage{wrapfig}
\usepackage{graphicx}

\mdfdefinestyle{mystyle}{%
  rightline=true,
  innerleftmargin=10,
  innerrightmargin=10,
  outerlinewidth=3pt,
  topline=false,
  rightline=true,
  bottomline=false,
  skipabove=\topsep,
  skipbelow=\topsep
}

\newtcolorbox{myboxi}[1][]{
  breakable,
  title=#1,
  colback=red!5,
  colbacktitle=red!5,
  coltitle=black,
  fonttitle=\bfseries,
  bottomrule=0pt,
  toprule=0pt,
  leftrule=2pt,
  rightrule=2pt,
  titlerule=0pt,
  arc=0pt,
  outer arc=0pt,
  colframe=red,
}

\newtcolorbox{myboxnote}[1][]{
  breakable,
  title=#1,
  colback=orange!0,
  colbacktitle=orange!0,
  coltitle=black,
  fonttitle=\bfseries,
  bottomrule=0pt,
  toprule=0pt,
  leftrule=2pt,
  rightrule=2pt,
  titlerule=0pt,
  arc=0pt,
  outer arc=0pt,
  colframe=orange,
}

\newtcolorbox{myboxii}[1][]{
  breakable,
  freelance,
  title=#1,
  colback=white,
  colbacktitle=white,
  coltitle=black,
  fonttitle=\bfseries,
  bottomrule=0pt,
  boxrule=0pt,
  colframe=white,
  overlay unbroken and first={
  \draw[red!75!black,line width=3pt]
    ([xshift=5pt]frame.north west) -- 
    (frame.north west) -- 
    (frame.south west);
  \draw[red!75!black,line width=3pt]
    ([xshift=-5pt]frame.north east) -- 
    (frame.north east) -- 
    (frame.south east);
  },
  overlay unbroken app={
  \draw[red!75!black,line width=3pt,line cap=rect]
    (frame.south west) -- 
    ([xshift=5pt]frame.south west);
  \draw[red!75!black,line width=3pt,line cap=rect]
    (frame.south east) -- 
    ([xshift=-5pt]frame.south east);
  },
  overlay middle and last={
  \draw[red!75!black,line width=3pt]
    (frame.north west) -- 
    (frame.south west);
  \draw[red!75!black,line width=3pt]
    (frame.north east) -- 
    (frame.south east);
  },
  overlay last app={
  \draw[red!75!black,line width=3pt,line cap=rect]
    (frame.south west) --
    ([xshift=5pt]frame.south west);
  \draw[red!75!black,line width=3pt,line cap=rect]
    (frame.south east) --
    ([xshift=-5pt]frame.south east);
  },
}

\usepackage{fontawesome5}
\usepackage{fancyhdr} 
\usepackage{blindtext} 
\usepackage{makecell}
\usepackage{framed}

\DeclareCaptionFont{black}{\color{black}}

\definecolor{myblue}{rgb}{0.9, 0.1, 0.94}
\definecolor{mygreen}{rgb}{0.64, 0.56, 0.88}
\definecolor{myyellow}{rgb}{0.68, 0.6, 0.1}
\definecolor{fancygreen}{rgb}{0.33, 0.68, 0.20}
\definecolor{salmon}{rgb}{0.94, 0.52, 0.49}
\definecolor{tablegreen}{rgb}{0.82, 0.94, 0.75}
\definecolor{tableblue}{rgb}{0.81, 0.90, 0.94}
\definecolor{tablered}{rgb}{0.97, 0.85, 0.85}
\definecolor{tableorange}{rgb}{0.96, 0.85, 0.81}

\newenvironment{itemize*}%
 {\leftmargini=10pt\begin{itemize}%
  \setlength{\itemsep}{0pt}%
  \setlength{\parskip}{0pt}%
  }%
 {\end{itemize}}
\newenvironment{enumerate*}%
 {\begin{enumerate}%
  \setlength{\itemsep}{0pt}%
  \setlength{\parskip}{0pt}}%
 {\end{enumerate}}

\usepackage{xcolor}
\usepackage{listings}  

\newcommand\JSONnumbervaluestyle{\color{blue}}
\newcommand\JSONstringvaluestyle{\color{red}}

\newif\ifcolonfoundonthisline

\makeatletter

\lstdefinestyle{json}
{
  showstringspaces    = false,
  keywords            = {false,true},
  alsoletter          = 0123456789.,
  morestring          = [s]{"}{"},
  stringstyle         = \ifcolonfoundonthisline\JSONstringvaluestyle\fi,
  MoreSelectCharTable =%
    \lst@DefSaveDef{`:}\colon@json{\processColon@json},
  basicstyle          = \ttfamily,
  keywordstyle        = \ttfamily\bfseries,
}

\newcommand\processColon@json{%
  \colon@json%
  \ifnum\lst@mode=\lst@Pmode%
    \global\colonfoundonthislinetrue%
  \fi
}

\lst@AddToHook{Output}{%
  \ifcolonfoundonthisline%
    \ifnum\lst@mode=\lst@Pmode%
      \def\lst@thestyle{\JSONnumbervaluestyle}%
    \fi
  \fi
  \lsthk@DetectKeywords%
}

\lst@AddToHook{EOL}%
  {\global\colonfoundonthislinefalse}

\makeatother

\usepackage{etoolbox}
\usepackage{natbib}
\usepackage{url}
\newcounter{bibcount}
\makeatletter
\patchcmd{\@lbibitem}{\item[}{\item[\hfil\stepcounter{bibcount}{[\thebibcount]}}{}{}
\renewcommand\NAT@bibsetup%
  [1]{\setlength{\leftmargin}{\bibhang}\setlength{\itemindent}{-\parindent}%
      \setlength{\itemsep}{\bibsep}\setlength{\parsep}{\z@}}
\makeatother

\definecolor{mybrown}{RGB}{128,64,0}

\definecolor{titlecolor}{HTML}{4c9cff}
\definecolor{coolblue3}{rgb}{0.91, 0.94, 0.98}

\begin{document}

\title{Speed by Simplicity: A Single-Stream Architecture for \\ Fast Audio-Video Generative Foundation Model}

\fancypagestyle{firstpage}{
  \fancyhf{}
  \rhead{\raisebox{-0.9cm}{\includegraphics[height=0.7cm]{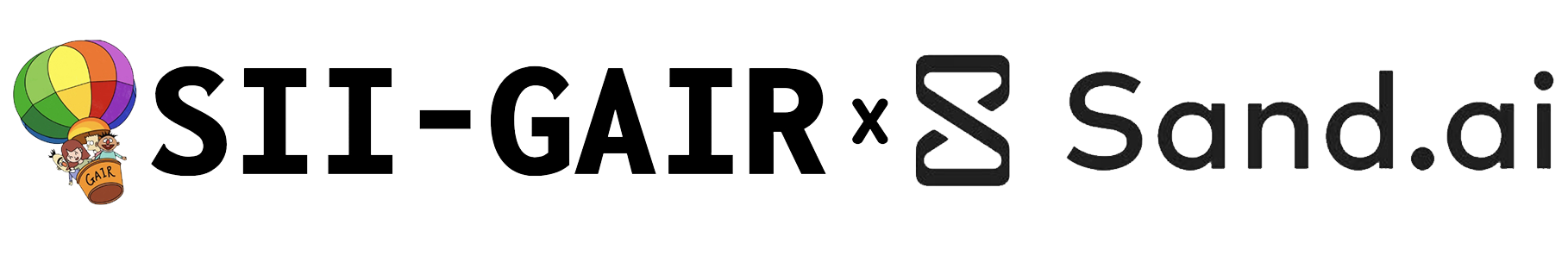}}}
  \cfoot{\thepage}
  \renewcommand{\headrulewidth}{0pt}
  \renewcommand{\headrule}{}
}

\author{\textbf{SII-GAIR\thanks{Corresponding author: Pengfei Liu $\langle$pengfei@sjtu.edu.cn$\rangle$.} \quad \& \quad Sand.ai\thanks{Corresponding author: Yue Cao $\langle$caoyue@sand.ai$\rangle$.}}}
\maketitle
\thispagestyle{firstpage}
\rhead{
\raisebox{-0.2cm}{\includegraphics[height=0.7cm]{assets/logo.png}}
}
\setlength{\headsep}{2mm} 
\renewcommand{\thefootnote}{}
\vspace{-15pt}
\begin{center}
    \quad \textbf{Open Source:} 
\quad \href{https://github.com/GAIR-NLP/daVinci-MagiHuman}{\textcolor{black}\faGithub\ Code}
\quad \href{https://huggingface.co/GAIR/daVinci-MagiHuman}{\raisebox{-.15em}{\includegraphics[height=1em]{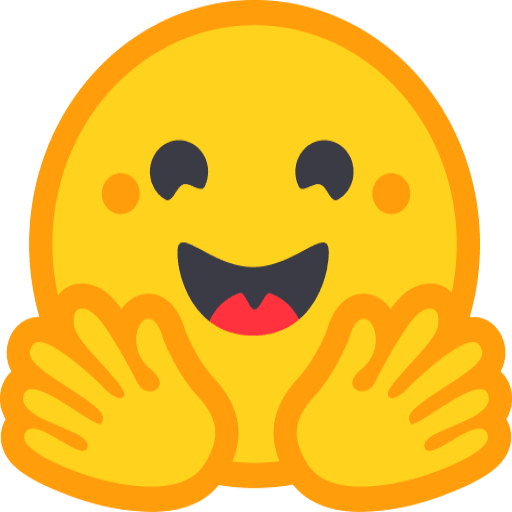}}\ Models}
\quad \href{https://huggingface.co/spaces/SII-GAIR/daVinci-MagiHuman}{\textcolor{violet}\faDatabase\ Demo}
\end{center}

\vspace{10pt}

\begin{abstract}
We present \textbf{daVinci-MagiHuman}, an open-source audio-video generative foundation model for human-centric generation. daVinci-MagiHuman jointly generates synchronized video and audio using a single-stream Transformer that processes text, video, and audio within a unified token sequence via self-attention only. This single-stream design avoids the complexity of multi-stream or cross-attention architectures while remaining easy to optimize with standard training and inference infrastructure. The model is particularly strong in human-centric scenarios, producing expressive facial performance, natural speech-expression coordination, realistic body motion, and precise audio-video synchronization. It supports multilingual spoken generation across Chinese (Mandarin and Cantonese), English, Japanese, Korean, German, and French. For efficient inference, we combine the single-stream backbone with model distillation, latent-space super-resolution, and a Turbo VAE decoder, enabling generation of a 5-second 256p video in 2 seconds on a single H100 GPU. In automatic evaluation, daVinci-MagiHuman achieves the highest visual quality and text alignment among leading open models, along with the lowest word error rate (14.60\%) for speech intelligibility. In pairwise human evaluation, it achieves win rates of 80.0\% against Ovi~1.1 and 60.9\% against LTX~2.3 over 2{,}000 comparisons. We open-source the complete model stack, including the base model, the distilled model, the super-resolution model, and the inference codebase.
\end{abstract}

\begin{figure}[h!]
    \centering
    \includegraphics[width=1.0\textwidth]{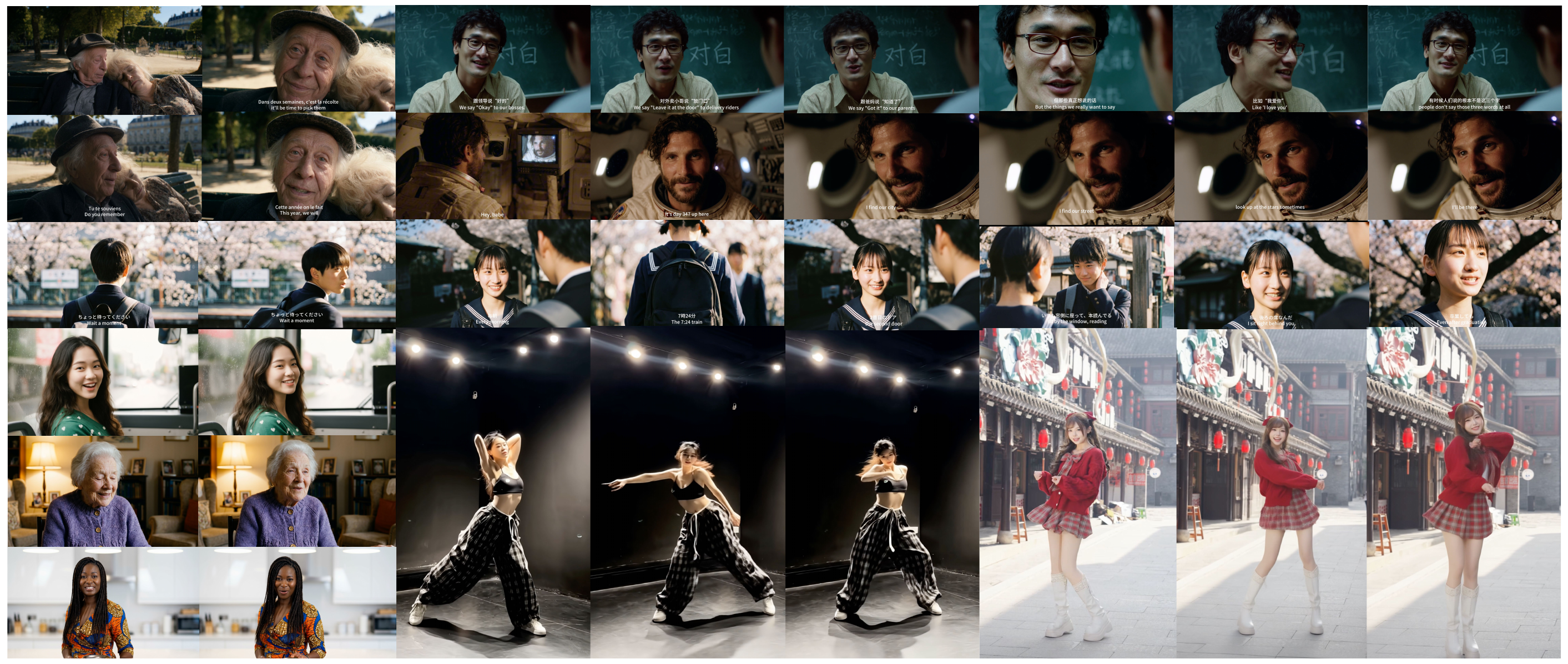}
    \caption{\textbf{Examples of videos generated by daVinci-MagiHuman}}
    \label{fig:main}
\end{figure}

\pagestyle{fancy}
\lhead{\rightmark}
\renewcommand{\headrulewidth}{0.7pt}
\setlength{\headsep}{5mm}

\clearpage

\newpage

\renewcommand{\thefootnote}{\arabic{footnote}}
\setcounter{footnote}{0}  

\input{1_overview}
\input{2_method}
\input{3_eval}

\newpage
\bibliographystyle{acl_natbib}
\bibliography{main}
\clearpage

\appendix
\section{Authors}

The authors are listed in alphabetical order, excluding the project leaders.

\vspace{1em}
\begin{multicols}{3} 
\noindent
Ethan Chern \\
Hansi Teng \\
Hanwen Sun \\
Hao Wang \\
Hong Pan \\
Hongyu Jia \\
Jiadi Su \\
Jin Li \\
Junjie Yu \\
Lijie Liu \\
Lingzhi Li \\
Lyumanshan Ye \\
Min Hu \\
Pengfei Liu \\
Qiangang Wang \\
Quanwei Qi \\
Steffi Chern \\
Tao Bu \\
Taoran Wang \\
Teren Xu \\
Tianning Zhang \\
Tiantian Mi \\
Weixian Xu \\
Wenqiang Zhang \\
Wentai Zhang \\
Xianping Yi \\
Xiaojie Cai \\
Xiaoyang Kang \\
Yan Ma \\
Yixiu Liu \\
Yue Cao \\
Yunbo Zhang \\
Yunpeng Huang \\
Yutong Lin \\
Zewei Tao \\
Zhaoliang Liu \\
Zheng Zhang \\
Zhiyao Cen \\
Zhixuan Yu \\
Zhongshu Wang \\
Zhulin Hu \\
Zijin Zhou \\
Zinan Guo 
\end{multicols}

\noindent{\textbf{Project Leader}}\\
Yue Cao, Pengfei Liu
\end{document}

%% file: 1_overview.tex
\section{Introduction}
Video generation has advanced rapidly in recent years, and the frontier is now shifting from silent video synthesis to the joint generation of synchronized video and audio. Although closed-source models such as Veo 3~\citep{veo3}, Sora 2~\citep{openai2025sora2}, and Kling 3.0~\citep{kuaishou2026kling3} have shown impressive capabilities, open-source progress (\textit{e.g.} Ovi~\citep{low2025ovi}, LTX-2~\citep{hacohen2026ltx}) in this direction remains limited. In particular, it remains challenging to build an open model that combines strong generation quality, multilingual support, and inference efficiency with a simple and scalable architecture.

In this report, we present \textbf{daVinci-MagiHuman}, an open-source audio-video generation model built to address these challenges. While leading open-source models ~\cite{hacohen2026ltx, wan2025wan, low2025ovi, team2026mova} typically rely on heavily specialized multi-stream designs, our model adopts a single-stream Transformer that models text, video, and audio within a shared-weight backbone. This design is simple at the model architectural level and easy to optimize jointly with training and inference infrastructure, making it better suited for future research and community development.

daVinci-MagiHuman is particularly strong in human-centric generation. The model performs especially well in scenarios that require expressive character acting, natural coordination between voice and facial expression, realistic body movement, and accurate audio-video synchronization. It also generalizes well across languages, delivering strong performance in Chinese (Mandarin and Cantonese), English, Japanese, Korean, German, and French, with support for additional languages beyond these major ones.

daVinci-MagiHuman is also designed for fast inference. The single-stream architecture is both hardware-friendly and infrastructure-friendly, making inference optimization significantly easier. In addition, we accelerate generation with latent-space super-resolution, which reduces the computation required for high-resolution video generation. As a result, our distilled model can generate a 5-second 256p video in 2 seconds and a 5-second 1080p video in 38 seconds on a single H100 GPU. These results make the model suitable not only for offline content creation, but also for latency-sensitive interactive applications.

To support future research and development, we fully open-source the complete model stack, including the base model, the distilled model, the super-resolution model, and the inference codebase. We hope this release can provide the community with a practical and extensible foundation for future work on audio-video generation.

Overall, daVinci-MagiHuman provides a strong open-source foundation for audio-video generation by combining architectural simplicity, strong human-centric quality, multilingual capability, and fast inference. Its main highlights are as follows:

\paragraph{Simple single-stream architecture}
A single-stream Transformer for text, video, and audio, avoiding the complexity of heavily specialized multi-stream architectures while remaining easy to optimize together with training and inference infrastructure.

\paragraph{Strong human-centric generation quality}
Particularly strong results in expressive human generation, including natural emotion, speech-expression coordination, facial performance, body motion, and audio-video synchronization.

\paragraph{Broad multilingual capability}
Strong spoken audio-video generation across multiple languages, including Chinese (Mandarin and Cantonese), English, Japanese, Korean, German, and French, with support for additional languages beyond these major ones.

\paragraph{Fast inference}
Efficient generation enabled by the single-stream backbone, latent-space super-resolution, and inference-level optimization.

\paragraph{Fully open-source release}
The complete model stack is released, including the diffusion model, the super-resolution model, and the inference codebase.

%% file: 2_method.tex
\section{Methodology}
\begin{figure}[htbp]
    \centering
    \includegraphics[width=0.9\linewidth]{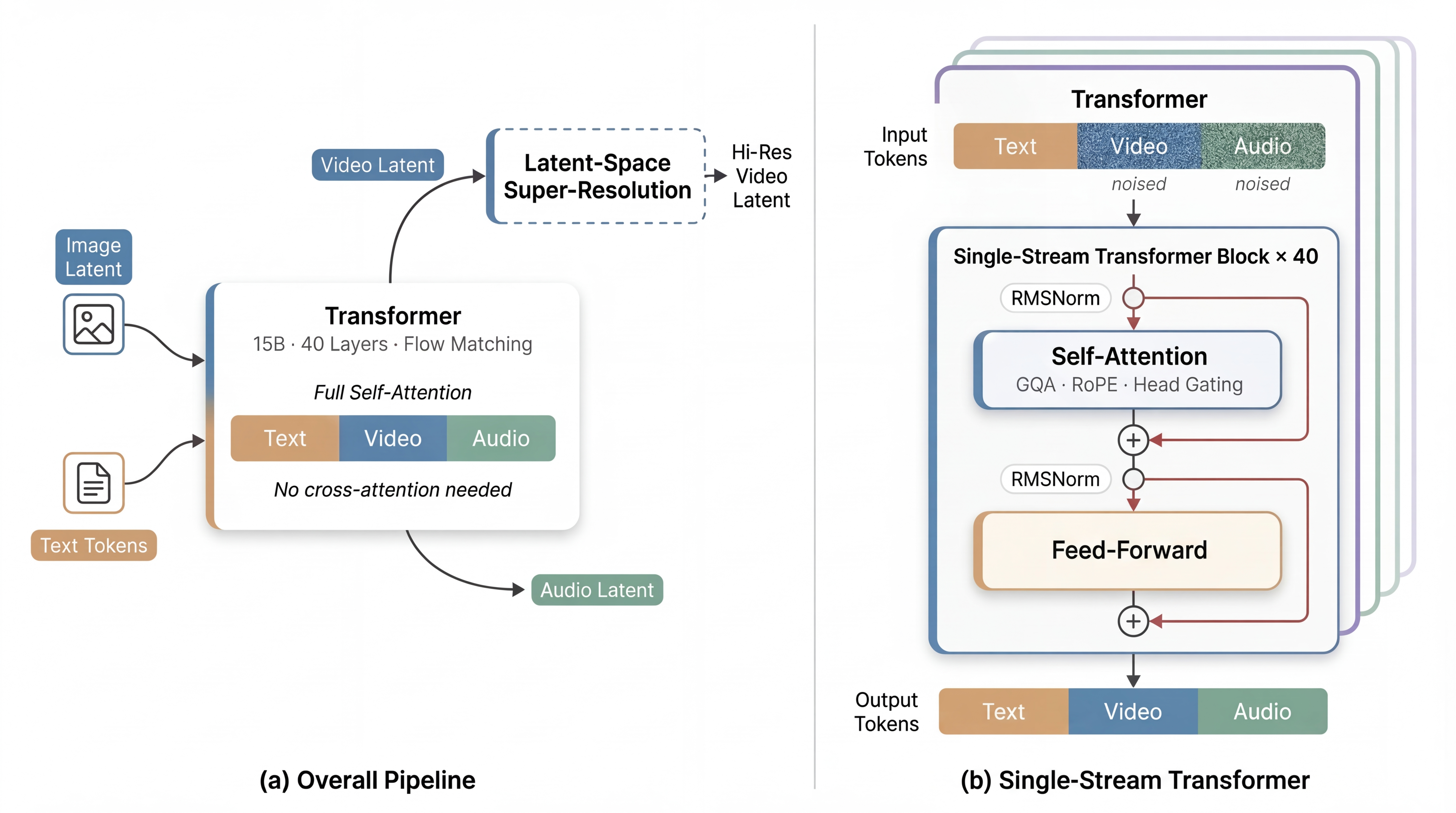}
    \caption{\textbf{Overall architecture.} (a) The base generator takes text tokens, a reference image latent, and noisy video and audio tokens as input, and jointly denoises the video and audio tokens with a single-stream Transformer. All modalities are processed within a unified token sequence using self-attention only, without separate cross-attention or fusion modules. A latent-space super-resolution stage can be further applied to refine the generated video at higher output resolutions. (b) The single-stream Transformer adopts a sandwich architecture layout: the first and last 4 layers use modality-specific projections and normalization parameters, while the middle 32 layers share the main Transformer parameters across modalities. Each block uses per-head gating in attention, and the model contains no explicit timestep embedding.}

    \label{fig:architecture}
\end{figure}

daVinci-MagiHuman is designed to balance architectural simplicity, strong generation quality, and fast inference. To achieve this, we build the model around several key design choices. In this section, we describe the main techniques behind the system.

\paragraph{Single-Stream Transformer}
Recent open-source video generation models~\cite{wan2025wan, kong2024hunyuanvideo} commonly adopt dual-stream architectures, where tokens from text and video are processed by partially separate branches and fused through cross-attention or other dedicated modules.

In audio-video generation~\cite{hacohen2026ltx, low2025ovi}, this design trend becomes even stronger, since the model must handle video and audio signals with different temporal structures and semantic patterns. As a result, many models adopt separate pathways for video and audio, dedicated fusion blocks, or modality-specific alignment modules. While such designs can be useful, they also make the overall architecture substantially more complex. This added complexity creates practical challenges beyond model design: multi-stream architectures introduce more irregular computation patterns, making implementation and optimization much harder in practice.

To address these issues, we adopt a single-stream Transformer architecture. Instead of maintaining separate pathways for different modalities, we represent text, video, and audio tokens within a shared backbone and model them using a unified stack of self-attention layers. This design keeps the architecture simple, reduces engineering complexity, and is easier to optimize jointly at both the model and infrastructure levels.

Figure~\ref{fig:architecture}(b) illustrates the core architecture. Our model uses a 15B-parameter, 40-layer single-stream Transformer backbone that jointly denoises video and audio at every step. Several design choices are central to its simplicity and effectiveness:

\begin{itemize}
    \item \textbf{Sandwich Architecture Layout.} The 40-layer Transformer is not fully homogeneous. The first and last 4 layers use modality-specific projections and RMSNorm parameters, while the middle 32 layers share the main Transformer parameters across modalities. This sandwich-style layout preserves modality-sensitive processing near the input and output boundaries while keeping most computation in a common representation space for deep multimodal fusion.
    \item \textbf{Timestep-Free Denoising.} Unlike original DiT architectures~\citep{peebles2023scalable} that inject diffusion timestep information through explicit timestep embeddings or AdaLN conditioning, our denoiser contains no dedicated timestep pathway. Following recent observations in ~\citep{sun2025noise, tang2025exploring}, the model receives the current noisy video and audio latents and infers the denoising state directly from the inputs themselves.
    \item \textbf{Per-Head Gating.} In each attention block, we follow the recent practice in large language models (LLMs)~\cite{qiu2025gated} of introducing an additional scalar gate for every attention head and use a sigmoid to modulate the attention output before the output projection. Concretely, if $o_h$ denotes the output of the $h$-th attention head and $g_h$ is the corresponding learned gate, the gated output is
    \[
    \tilde{o}_h = \sigma(g_h) \, o_h,
    \]
    This mechanism is introduced to improve numerical stability during training and to enhance model representability, while adding only minimal architectural overhead.
    \item \textbf{Unified Conditioning Without Extra Modules.} We handle denoising and reference signals with a minimal unified interface rather than introducing dedicated conditioning branches. Denoising video and audio tokens, together with text and optional image conditions, are all represented in the same latent/token space and processed by the same model. This design allows us to support multiple conditioning and generation settings with a simple shared architecture instead of task-specific fusion modules.
\end{itemize}

\paragraph{Efficient Inference Techniques}
Beyond the single-stream backbone itself, we improve inference efficiency with several complementary techniques: latent-space super-resolution, a turbo video decoder, full-graph compilation and distillation.
\begin{itemize}
    \item \textbf{Latent-Space Super-Resolution.} Directly generating high-resolution video from scratch remains expensive because the video token count grows quickly with spatial resolution. To reduce this cost, we adopt a two-stage pipeline: the base model first generates video and audio latents at a lower base resolution, and a super-resolution stage then refines the result at higher resolution. We perform this refinement in latent space rather than pixel space because it stays aligned with the native diffusion representation, reuses the same overall backbone architecture, and avoids an extra VAE decode-and-encode round trip. Concretely, we upsample the video latent with trilinear interpolation, inject additional noise, and refine it with only 5 extra denoising steps using a dedicated super-resolution checkpoint. In the 1080p setting, the super-resolution model additionally enables local attention in many layers to control high-resolution attention cost. Although the stage is primarily designed to improve the video output, it still takes audio latent tokens as input and predicts video and audio jointly within the same backbone. In practice, only the video latent is explicitly updated during the super-resolution sampling step, while the audio latent from the base stage is reused in a noised form as auxiliary input. This design keeps the refinement process coupled to the audio signal, which is especially useful when the base-resolution video is very coarse and lip synchronization would otherwise be harder to preserve.
    
    \item \textbf{Turbo VAE Decoder.} We use Wan2.2 VAE~\citep{wanteam2025wan2.2} for encoding because of its high spatial-temporal compression ratio, while replacing the original video decoder at inference time with a lightweight re-trained Turbo VAE decoder~\cite{zou2026turbo}. This substantially reduces decoding overhead and is important because decoding lies on the critical path of both the base generator and the super-resolution pipeline.
    
    \item \textbf{Full-Graph Compilation.} We further integrate \href{https://github.com/SandAI-org/MagiCompiler}{MagiCompiler}, our full-graph PyTorch compiler, into the inference stack. By fusing operators across Transformer layer boundaries and consolidating distributed communication into fewer collective calls, it provides around 1.2$\times$ speedup on H100.

    \item \textbf{Distillation.} To reduce cost in inference, we apply DMD-2~\cite{yin2024improved} to distill the base generator. As a result, the distilled model can generate with only 8 denoising steps without CFG, while maintaining strong generation quality. Unless otherwise specified, the latency numbers reported in Section~\ref{sec:system} use this distilled model.

\end{itemize}

%% file: 3_eval.tex
\section{Evaluation}
\label{sec:system}

We compare daVinci-MagiHuman with two leading open-source baselines, Ovi~1.1~\cite{low2025ovi} and LTX~2.3~\cite{hacohen2026ltx}. Our evaluation covers three aspects: automatic quality metrics, pairwise human preference, and inference efficiency.

\begin{table}[htbp]
    \centering
    \footnotesize
    \caption{Quantitative Analysis of Ovi-1.1, LTX-2.3, and daVinci-MagiHuman.}
    \label{tab:video_scores}
    \setlength{\tabcolsep}{10pt}
    \renewcommand{\arraystretch}{1.3}
    \resizebox{\textwidth}{!}{%
    \begin{tabular}{@{} l ccc c @{}}
    \toprule
     & \multicolumn{3}{c}{\textsc{Video Quality Score}} & \textsc{Audio Quality Score} \\
    \cmidrule(lr){2-4} \cmidrule(lr){5-5}
    \textbf{Model}
     & \textit{Visual Quality\,$\uparrow$}
     & \textit{Text Alignment\,$\uparrow$}
     & \textit{Physical Consistency\,$\uparrow$}
     & \textit{WER\,$\downarrow$} \\
    \midrule
    OVI 1.1               & 4.73 & 4.10 & 4.41 & 40.45\% \\
    LTX 2.3               & 4.76 & 4.12 & \textbf{4.56} & 19.23\% \\
    \textbf{daVinci-MagiHuman}      & \textbf{4.80} & \textbf{4.18} & 4.52 & \textbf{14.60\%} \\
    \bottomrule
    \end{tabular}%
    }
\end{table}

\paragraph{Quantitative Quality Benchmark}
We first report quantitative quality results against Ovi~1.1 and LTX~2.3. For video quality, we evaluate on VerseBench~\cite{wang2025universe} and adopt VideoScore2~\cite{he2025videoscore2} to measure visual quality, text alignment, and physical consistency. For audio quality, we evaluate speech intelligibility on TalkVid-Bench~\citep{chen2025talkvidlargescalediversifieddataset} using word error rate (WER), where lower is better. All generated audio is transcribed by GLM-ASR~\citep{zai2025glmasr}. For CJK languages, we compute WER at the character level to avoid inconsistencies from word segmentation.

As shown in Table~\ref{tab:video_scores}, daVinci-MagiHuman achieves the best visual quality and text alignment scores among the compared models, while also obtaining the lowest WER of 14.60\%. This substantially outperforms Ovi~1.1 (40.45\%) and also improves over LTX~2.3 (19.23\%). LTX~2.3 performs best on physical consistency, but daVinci-MagiHuman remains competitive on this metric and achieves the strongest overall balance across visual and audio quality.

\begin{table}[htbp]
    \centering
    \caption{Time breakdown (in seconds) for generating a 5-second video at different resolutions. All results are measured on a single H100 GPU. The base stage uses the distilled model, and all decode times are measured with the Turbo VAE decoder.}
    \label{tab:time_breakdown}
    \setlength{\tabcolsep}{8pt}
    \renewcommand{\arraystretch}{1.3}
    \small
    \begin{tabular}{@{} l cccc @{}}
    \toprule
    \textbf{Resolution}
     & \textit{Base}
     & \textit{SR}
     & \textit{Decode}
     & \textit{Total} \\
    \midrule
    \textsc{256p} & 1.6 & -- & 0.4 & 2.0 \\
    \textsc{540p} & 1.6 & 5.1 & 1.3 & 8.0 \\
    \textsc{1080p} & 1.6 & 31.0 & 5.8 & 38.4 \\
    \bottomrule
    \end{tabular}
\end{table}

\paragraph{Human Evaluation}
We further conduct a pairwise human evaluation against two open-source audio-video models: Ovi~1.1~\cite{low2025ovi} and LTX~2.3~\cite{hacohen2026ltx}. A total of 10 human raters each judge 200 randomized pairs, including 100 comparisons against each competitor, for a total of 2{,}000 comparisons. Raters select the preferred clip or declare a tie based on overall audio-video quality, synchronization, and naturalness.

\begin{figure}[htbp]
    \centering
    \includegraphics[width=0.8\linewidth]{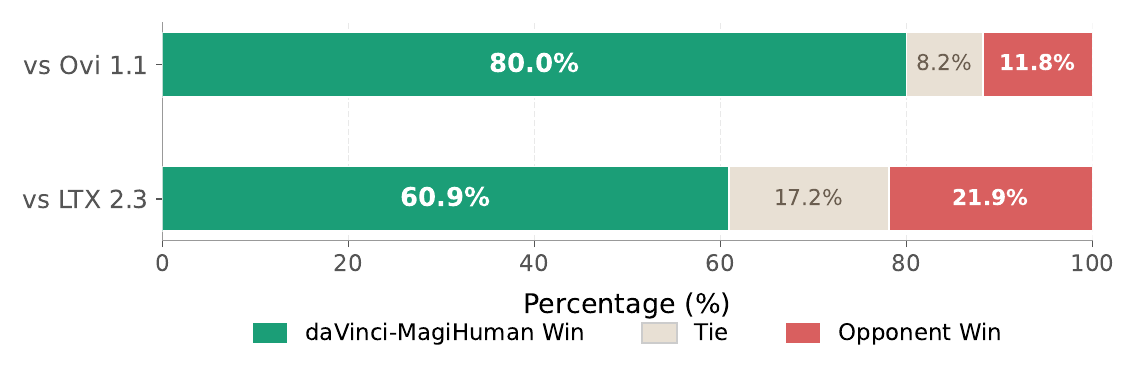}
    \caption{\textbf{Human evaluation results.} Pairwise preference rates of daVinci-MagiHuman against two open-source audio-video generation models, aggregated over 10 raters and 2{,}000 comparisons.}
    \label{fig:human_eval}
\end{figure}

As shown in Figure~\ref{fig:human_eval}, daVinci-MagiHuman is consistently preferred over both baselines, achieving win rates of 80.0\% against Ovi~1.1 and 60.9\% against LTX~2.3. The corresponding opponent win rates are 11.8\% and 21.9\%, with tie rates of 8.2\% and 17.2\%, respectively. Overall, these results indicate a clear human preference for daVinci-MagiHuman across the tested pairwise comparisons.

\paragraph{Inference Efficiency}
We finally evaluate inference efficiency from the end-to-end latency perspective. Table~\ref{tab:time_breakdown} provides a stage-wise breakdown on a single H100 GPU. In these measurements, the base stage always runs at 256p using the distilled model, while higher output resolutions are obtained through the super-resolution stage and decoding is performed with the Turbo VAE decoder. As a result, the base-stage latency remains constant across output resolutions, and the additional cost at higher resolutions is dominated by super-resolution and decoding. Even so, the entire pipeline takes only 38.4 seconds to produce a 5-second 1080p video.